\documentclass[journal]{./IEEEtran}
\usepackage{cite}
\usepackage[pdftex]{graphicx}
\graphicspath{{images/}}
\usepackage{amsmath}
\usepackage{array}
\ifCLASSOPTIONcompsoc
  \usepackage[size=normalsize,labelfont=sf,textfont=sf]{caption}
  \usepackage[size=normalsize,labelfont=sf,textfont=sf]{subcaption}
\else
  \usepackage[size=footnotesize]{caption}
  \usepackage[size=footnotesize]{subcaption}
\fi
\usepackage{stfloats}
\usepackage{url}
\usepackage{hyperref}       % hyperlinks
\usepackage{amsfonts}       % blackboard math symbols
\usepackage{nicefrac}       % compact symbols for 1/2, etc.
\usepackage{microtype}      % microtypography
\usepackage{mathtools}      % math utilities
\usepackage{bm}             % bold math
\usepackage{bbm}            % bold math digits
\usepackage{makecell}       % prettier tables
\usepackage{multirow}       % table multi-row cells

\newcommand{\R}{\mathbb{R}}
\newcommand{\etal}{\textit{et al. }}

\begin{document}
\title{A Graph Convolutional Network with Signal Phasing Information for Arterial Traffic Prediction}

\author{Victor~Chan,
        Qijian~Gan,
        and~Alexandre~Bayen% <-this % stops a space
% \thanks{M. Shell was with the Department
% of Electrical and Computer Engineering}% <-this % stops a space
% \thanks{Manuscript received April 19, 2005; revised August 26, 2015.}
}


\markboth{}%
{Chan, Gan, and Bayen: Graph Convolutional Networks with Signal Phasing Information for Arterial Traffic Flow Prediction}



\maketitle

\begin{abstract}
Accurate and reliable prediction of traffic measurements plays a crucial role in the development of modern intelligent transportation systems. Due to more complex road geometries and the presence of signal control, arterial traffic prediction is a level above freeway traffic prediction. Many existing studies on arterial traffic prediction only consider temporal measurements of flow and occupancy from loop sensors and neglect the rich spatial relationships between upstream and downstream detectors. As a result, they often suffer large prediction errors, especially for long horizons. We fill this gap by enhancing a deep learning approach, Diffusion Convolutional Recurrent Neural Network, with spatial information generated from signal timing plans at targeted intersections. Traffic at signalized intersections is modeled as a diffusion process with a transition matrix constructed from the phase splits of the signal phase timing plan. We apply this novel method to predict traffic flow from loop sensor measurements and signal timing plans at an arterial intersection in Arcadia, CA. We demonstrate that our proposed method yields superior forecasts; for a prediction horizon of 30 minutes, we cut the MAPE down to 16\% for morning peaks, 10\% for off peaks, and even 8\% for afternoon peaks. In addition, we exemplify the robustness of our model through a number of experiments with various settings in detector coverage, detector type, and data quality.
\end{abstract}


% \fi
\IEEEpeerreviewmaketitle

\section{Introduction}

\label{introduction}

\IEEEPARstart{T}{he} problem of efficient transportation has typically been a hardware and civil engineering problem, as companies have developed faster and cleaner cars, built carefully-designed freeways, and architected roads in cities. With the rise of intelligent transportation systems (ITS), the problem has shifted focus to the fields of mathematics, statistics, and computer science. As governments install more sensors in road networks and collect ever-increasing amounts of data, research has begun to concentrate on designing improved prediction and control techniques. Traffic flow and speed prediction has numerous applications, such as freeway metering, travel time prediction, intelligent intersection signal control, and traffic simulation software. With accurate traffic flow forecasts, cities can better plan logistics and allocation of resources for construction, road development, and safety. Predictions can also be leveraged to optimize signal control at intersections, saving commuters valuable time and reducing consumption of gas and electricity.

Historically, a wide variety of models have been used for traffic flow prediction. Although they are often grouped into parametric and nonparametric categories, or classified as statistics or machine learning, most models are closely-related and have overlapping properties \cite{comparison}. Statistical methods such as Kalman filters \cite{kalman_filtering}, exponential smoothing \cite{exponential_smoothing}, and techniques in the ARMA family \cite{ARIMA_OG, ARIMA, ARIMAX, SARIMA, ARMAX} typically rely on strong priors and assumptions about data distributions; as a result, traffic experts must carefully select and structure the models. Because of this, parametric methods present a trade-off between easy interpretation and practicality \cite{parametric_comparison}. Nonparametric methods, more flexible and expressive than parametric models, have surged in popularity as hardware has been upgraded and more powerful algorithms have been developed. Methods prevalent in the literature include nearest neighbors regression \cite{KNN_Similarity_Degree}, principal component analysis \cite{lowrank}, decision trees, support vector machines \cite{etaSVR}, and fuzzy rule-based systems \cite{FRBS}.

Even these machine learning methods have been overshadowed by the rise of neural networks. Deep learning has gained much traction in recent years as data has become more readily available and computer power has exponentiated. Simple feed-forward neural networks have evolved into convolutional neural networks, long short-term memory, and graph convolutions. State-of-the-art algorithms utilize meta-learning and distillation \cite{MetaST, ST-MetaNet}, residual connections \cite{ST-ResNet}, attention \cite{complicated_cnn_lstm}, and adversarial training \cite{GCGAN, TrafficGAN}. Deep architectures open the door for a new generation of nonparametric models that are constantly improving prediction accuracy.

Overall, most prediction methods are very proficient at forecasting freeway data. Freeways are a mostly-closed system, with leakages only from on-ramps and off-ramps. Traffic flows smoothly from one sensor to the next with few interruptions; thus, freeway traffic data is typically smooth and clean. In contrast, arterial traffic is much noisier and more difficult to predict. At intersections, traffic signals and stop signs introduce exogenous factors that affect the speed and movement of cars. Moreover, elements such as pedestrians, bikes, parking, and driveways further complicate traffic patterns. Much existing literature focuses on freeway traffic prediction, but less work explores its arterial counterpart.

One strategy that has proved useful in overall traffic flow prediction is the graph convolution, which applies to the setting of predicting a label for a graph, given a set of graphs with their associated labels \cite{DCNN}. In the most general case, the graphs are directed and weighted, and the labels can be associated with any part of the graph, including the nodes, edges, and the graphs themselves. The spatial information from graph convolutions is significant in arterial traffic flow prediction because the detector graph is much more complicated than that of freeways. Graph convolutions have spatial structure built into the architecture, so they naturally account for the spatial relationships between detectors when predicting traffic flow.

Another consideration is the inclusion of different types of data as input for prediction. Most models treat the data as a time series, thus relying only on historical values of the data to forecast future values. Sometimes, extra features such as date, time, day of week, and exogenous events are included \cite{RSLDS}. We employ signal phase timing data from the traffic signals at our study site. Previously, signal phase timing data has been combined with detailed traffic knowledge to develop a system of equations to predict traffic flow \cite{signal_timing}. However, the model only applied to very short-term predictions, as it was intended for real-time signal control.

In this study, we focus on the Diffusion Convolutional Recurrent Neural Network (DCRNN) \cite{DCRNN}. We apply DCRNN to predict arterial traffic flow for detectors with full coverage. In order to adapt DCRNN to arterial traffic, we use novel signal phase timing data to construct the weighted transition matrix of the graph. Instead of modeling transition probabilities with road distances, which are not suitable for intersections, we calculate the phase split fraction from the phase split and cycle length. We demonstrate that using signal phase timing information reduces prediction error, especially for long horizon predictions. Moreover, we find through many ablation studies that the model does indeed learn the relationships between the detectors in the network.\footnote{Code available at \url{https://github.com/victorchan314/arterial_traffic_flow_predictor}}

The rest of this paper is organized as follows. In section \ref{literature_review}, we summarize current literature on traffic flow prediction, especially in deep learning. In section \ref{method}, we present the model we use and our strategy to append signal phase timing data. In section \ref{studysite_dataset}, we introduce the study site and dataset used in this report. In section \ref{analysis}, we analyze our arterial traffic flow forecasts and evaluate their effectiveness through many ablation studies. In section \ref{conclusion}, we draw conclusions based on our analysis.

%\IEEEpubidadjcol

\section{Related Work}

\label{literature_review}

\subsection{ARMA Models}

Of the numerous statistical methods for traffic flow prediction, we focus on models in the ARMA family, which have seen much success in general time series prediction. Although not the first, Ahmed and Cook applied an $\text{ARIMA}(0, 1, 3)$ model to forecast freeway occupancy and flow in 1979 \cite{ARIMA_OG}. Hamed, Al-Masaeid, and Said extended the model to predict arterial flow \cite{ARIMA}. Williams and Hoel showed that a weekly seasonal difference could make freeway flow stationary, thus cementing the theoretical justification for fitting ARMA models to traffic data \cite{SARIMA}. The field has been further expanded by the application of exogenous data to standard ARIMA models \cite{ARIMAX, ARMAX}.

\subsection{Deep Learning}

\paragraph{Recurrent Neural Networks}

Because traffic data is a time series, it makes sense to apply recurrent neural networks (RNN) to the prediction problem to learn temporal patterns. Long Short-Term Memory (LSTM) and Gated Recurrent Unit (GRU) architectures mitigate the vanishing gradient problem \cite{LSTM, LSTM_shallow, LSTM_GRU}. Zhao \etal input an origin-destination correlation matrix, which captures the correlation between different points in the detector network, to LSTM \cite{LSTM_ODC}. The above methods dealt with freeway data; in contrast, Mackenzie, Roddick, and Zito used a sparse distributed representation of data with binary vectors and showed that it was comparable to LSTM for arterial flow prediction \cite{HTM_LSTM}.

\paragraph{Recurrent Convolutional Neural Networks}

RNN methods simply include data from multiple sensors to extract the encoded spatial information, which is not very effective. To directly handle the spatial dimension, models evolved to synthesize RNNs with convolutional neural networks (CNN). Yu \etal filled an image of a road network with average link speeds and fed it sequentially into a 2D convolution and an LSTM to learn temporal relationships \cite{SRCN}. Yao \etal used start and end flow values for a two-channel image \cite{complicated_cnn_lstm}.

\paragraph{Graph Convolutions}

While CNNs consider spatial relationships in a more proper way, they still bear an inevitable mismatch with traffic data. Road networks are inherently graphs and not grids---they are not accurately represented as images. To this end, we turn to the graph convolution, which is perfect for traffic data. The graph structure is explicitly baked into the architecture of Graph Convolutional Networks (GCN) instead of being implicitly included with the data or imprecisely approximated with images.

Atwood and Towsley defined the Diffusion Convolutional Neural Network (DCNN), which uses the power series of the degree-normalized transition matrix to model a diffusion process; DCNN output high-quality results for citation graph datasets \cite{DCNN}. Li \etal adapted the DCNN with a seq2seq GRU architecture to create DCRNN \cite{DCRNN}. Other studies approximate the filters with a first order Chebyshev polynomial \cite{STGCN} or use a graph convolution for feature extraction \cite{T-GCN}. More recent works incorporate state-of-the-art innovations such as Wavenet \cite{Graph_WaveNet}, U-Net \cite{ST-UNet}, and attention \cite{ASTGCN, GSTNet} into GCNs. Fewer works take advantage of GCNs to forecast arterial data. Cui \etal used an adjacency matrix of nodes in a $k$-hop neighborhood to extract features of the graph before feeding them into an LSTM \cite{TGC-LSTM}. Guo \etal optimized the Laplace matrix in a graph convolution in GRU cells and showed that the learned matrices had high correlation with physical proximity \cite{OGCRNN}.

\paragraph{Other Deep Learning Methods}

There are many deep learning methods that have shown promise in traffic flow prediction, but do not leverage the graph convolution. Simpler works for predicting freeway flow utilize multilayer perceptrons \cite{genetic_NN}, stacked autoencoders \cite{SAE}, vector autoregression \cite{sparse_VAR}, CNNs \cite{images, CapsNet}, and ResNets \cite{ST-ResNet}. In order to reap the benefits from multiple models, Zhan \etal used a consensus ensemble system to prune outliers in arterial flow forecasts \cite{connected_corridors_consensus}.

The most recent works have incorporated elements of deep reinforcement learning and unsupervised learning into arterial flow prediction. They are often carefully assembled from complicated components that employ meta-learning \cite{MetaST, ST-MetaNet}, attention \cite{STANN}, and Generative Adversarial Networks (GAN) \cite{GCGAN, TrafficGAN} to learn feature representations of traffic. These methods provide flexible and expressive models that, if designed and trained properly, can easily outperform parametric and statistical methods. The additional parameters of these models also provide a way to incorporate extra data, such as signal phase timing information. There is still much to be explored and much room for improvement, especially with arterial traffic prediction.

\section{Method}

\label{method}

Most conventional traffic prediction methods only exploit temporal information. Spatial relations are ignored or not directly built into the architecture. Recently, a new method, DCRNN \cite{DCRNN}, has been proposed to directly integrate spatial information, such as sensor layouts, into the architecture. It models freeway traffic as a diffusion process where cars disperse from upstream sensors to downstream sensors. The parameters for the transition matrix rely on physical properties of the road network, such as distances between sensors, unlike other graph convolutions that use binary weights or learn the weights; thus, signal phase timing data is apt for the model. This spatio-temporal property of DCRNN has also been utilized in other applications and fields: travel time estimation \cite{travel_time_estimation}, ride-hailing demand \cite{ride_hailing}, air quality forecasting \cite{air_quality}, and distributed fleet control in reinforcement learning \cite{distributed_fleet_control}.

Different from the aforementioned studies, we apply DCRNN to arterial traffic prediction. We are one of the first studies to so; moreover, we are the first to use signal phase timing data with deep learning for arterial traffic prediction. We establish that it is possible to model traffic at adjacent arterial intersections as diffusion processes if the architecture is correctly constructed with the right parameter settings. A more detailed description of our model architecture is provided in the following subsections.

\subsection{DCRNN}

DCRNN relies on the title diffusion process to incorporate spatial information into the architecture. This is represented with a transition matrix between the network of sensors which, when multiplied with the state vector at time $t$, outputs the data point for time $t+1$. The transition matrix defines a \textit{diffusion convolution} operation that replaces matrix multiplications in a seq2seq RNN to comprise the DCRNN.

Let us define $D$ as the number of detectors in our network and $F$ as the number of features from each detector (flow, occupancy, etc.). Let us also define $H$ as the \textit{prediction horizon}, the number of time steps that we predict into the future, and $S$ as the \textit{window size}, the number of time steps we use to predict. Then each data point is an $X \in \R^{D \times F}$, and our goal is to learn a model that uses input $(X^{(t - S + 1)}, \hdots, X^{(t)})$ to predict $(X^{(t + 1)}, \hdots, X^{(t + H)})$.

We represent our system as a weighted directed graph $G = \{V, E, \bm{W}\}$, where the detectors are the vertices and the arterial roads are the edges. In our graph, $|V| = D$, and the transition matrix $\bm{W} \in \R^{D \times D}$ is a weighted adjacency matrix, with entry $\bm{W}_{i, j}$ representing the likelihood of transitioning from node $i$ to node $j$. These weights do not have to be probabilities and do not need to be normalized; they must simply be some function that is larger for nodes $j$ that are more likely destinations of cars from node $i$. We define $\bm{D_O} = \text{diag}(\bm{W}\bm{1})$ and $\bm{D_I} = \text{diag}(\bm{W^\top}{\bm{1}})$, where $\bm{1} \in \R^D$ is the all-ones vector and the $\text{diag}$ function takes in a vector and constructs a square matrix with the entries of the vector along its main diagonal. Thus, $\bm{D_O}, \bm{D_I} \in \R^{D \times D}$ are the normalization matrices for the forward and reverse diffusion processes, since traffic flow is affected by both upstream and downstream detectors.

These diffusion processes are represented as random walks on $G$ with a restart probability $\alpha \in [0, 1]$. Then the stationary distribution $\mathcal{P}$ of the forward diffusion process is
\begin{equation}
\mathcal{P} = \sum_{k=0}^\infty \alpha(1 - \alpha)^k(\bm{D_O}^{-1}\bm{W})^k
\end{equation}
The DCRNN model uses a truncated $K$-step diffusion process with learned weights for each step. The diffusion process, which we denote by $\bm{\mathfrak{F}}_\theta$, is parameterized by $\theta \in \R^{K \times 2}$ and acts on an input $X \in \R^{D \times F}$ to produce an output $Y \in \R^{D}$.
\begin{equation}
\begin{split}
\scalebox{0.95}{%
$\bm{\mathfrak{F}}_\theta(X; G, f) = \sum\limits_{k=0}^{K-1} \bigg(\theta_{k, 0}(\bm{D_O}^{-1}\bm{W})^k + \theta_{k, 1}(\bm{D_I}^{-1}\bm{W^\top})^k\bigg)X_{:, f}$
} \\
\text{for } f \in \{1, \hdots, F\}
\end{split}
\end{equation}

\begin{figure*}
  \centering
  \includegraphics[width=0.8\textwidth]{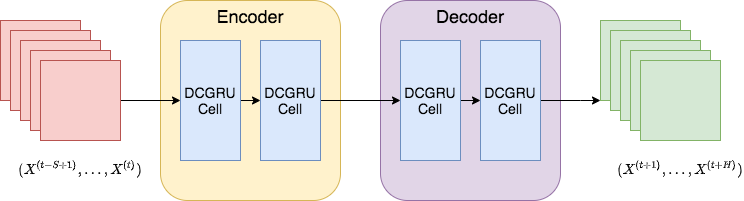}
  \caption{The seq2seq architecture of DCRNN. The encoder and decoder are both composed of multiple DCGRU cells.}
  \label{fig:dcrnn}
\end{figure*}

To incorporate diffusion convolutions into a model of the network, we use a Gated Recurrent Unit (GRU) \cite{GRU}, but with matrix multiplications replaced by the diffusion convolution. This constitutes the \textit{Diffusion Convolutional Gated Recurrent Unit} (DCGRU). Multiple DCGRUs are then stacked together in a seq2seq architecture, which finalizes the structure of DCRNN (Fig. \ref{fig:dcrnn}). In our paper, we use two cells in the encoder and two cells in the decoder. We feed in a sequence of $S$ inputs $X \in \R^{D \times F}$, and the next $H$ outputs (with earlier outputs recursively fed into the DCRNN to generate later outputs) are the predictions. The network is trained with backpropagation from loss incurred by our labeled data points. The authors also use scheduled sampling during training to switch between using ground truth labeled outputs and predictions from the DCRNN to generate later predictions.

Through the diffusion convolution and transition matrix, spatial information is baked into the architecture and not learned, allowing the model to learn parameters only for the relationships between the spatial and temporal information. Because of this embedded architecture, we need to train the model with a new transition matrix if detectors are added to or removed from the network. In order to adapt DCRNN for use with arteries, we modify the transition matrix.

\subsection{Transition Matrix}

The weights for the transition matrix in \cite{DCRNN} are derived from the road network distances between sensors. These distances are appropriate parameterizations for freeway traffic, as longer distances are correlated with slower diffusion rates. However, for an arterial intersection, road distance is inappropriate. Intersection roads are closely clustered, rendering any variation in distance insignificant.

Instead, we use \textit{phase split fraction}, defined as the fraction of a cycle during which cars are allowed to travel from the inbound sensor to the outbound sensor. The phase split fraction is calculated for each unique combination of intersection and plan. We use the fraction of time and not the actual number of seconds in order to normalize between busy intersections with longer cycles and smaller intersections with shorter cycles. In addition, because DCRNN assumes a static transition matrix, we use the planned phase split for the signal phase timing plan and not the actual phase split.

Let $P$ represent the set of phases active at a set of intersections of interest. For a phase $p \in P$, we let $p_{in}$ denote the inbound direction and $p_{out}$ denote the set of outbound directions of the phase. Let $d^{(i)}$ denote the $i$th detector in our dataset of $D$ detectors, $d_{dir}^{(i)}$ denote the direction of detector $i$, $I_{d^{(i)}}$ denote the intersection of detector $i$, and $\text{adj}(d^{(i)}, d^{(j)})$ be a boolean denoting whether detector $j$ is directly downstream from detector $i$, i.e. there is a direct path from detector $i$ to detector $j$. Let $L(I, p)$ denote the phase split of phase $p$ of intersection $I$. We compute the weights of the transition matrix as follows:
\begin{equation}
\bm{W}_{i, j} = \begin{cases}
\hfil 1 & \hspace*{-1.5cm} \text{if } I_{d^{(i)}} = I_{d^{(j)}}\\[0.2cm]
\dfrac{\splitfrac{\sum_{p \in P} \mathbbm{1}_{\text{adj}(d^{(i)}, d^{(j)})}\mathbbm{1}_{d_{dir}^{(i)} = p_{in}}\mathbbm{1}_{d_{dir}^{(j)} \in p_{out}}}{(L(I_{d^{(i)}}, p))}}{\frac{1}{2}\sum_{p \in P} L(I_{d^{(i)}}, p)} & \text{o.w.}
\end{cases}
\end{equation}
Because we use phase split fraction instead of road distances, we do not transform the weights with the Gaussian kernel as in \cite{DCRNN}. Instead, we leave the probabilities as the weights for the graph. As in \cite{DCRNN}, we zero out values in the matrix less than the threshold of $\varepsilon = 0.1$. In our transition matrices, we incorporate signal phase timing information for Through, Left Turn, and Right Turn directions. However, for both the upstream and downstream directions, we do not include U-Turns in our model. Overall, U-Turns contribute little flow to the data, especially during congested peak hours. In order to avoid this noise and not have to incorporate additional sensors in the opposite direction in our network, we ignore U-Turns.

\subsection{Flow Prediction}

In our study, we use two types of detectors: \textit{advance detectors}, placed in lanes about 100-200 feet before the intersection, and \textit{stopbar detectors}, located just before the intersections. Both types of detectors measure flow, occupancy, and speed. We use flow data in our model because most of the detectors in our network are advance detectors, for which flow measurements are the most reliable. In some cases, we include occupancy measurements during training in order to determine whether occupancy provides any benefit for flow prediction, but we disregard occupancy predictions, as the results are not as accurate as those of flow. Predicting flow instead of speed does not introduce any major changes to the methodology.

\section{Study Site and Dataset}

\label{studysite_dataset}

\subsection{Study Site}

\begin{figure}
  \centering
  \includegraphics[clip,trim={10.2cm 7.9cm 28.9cm 4.2cm},width=\columnwidth]{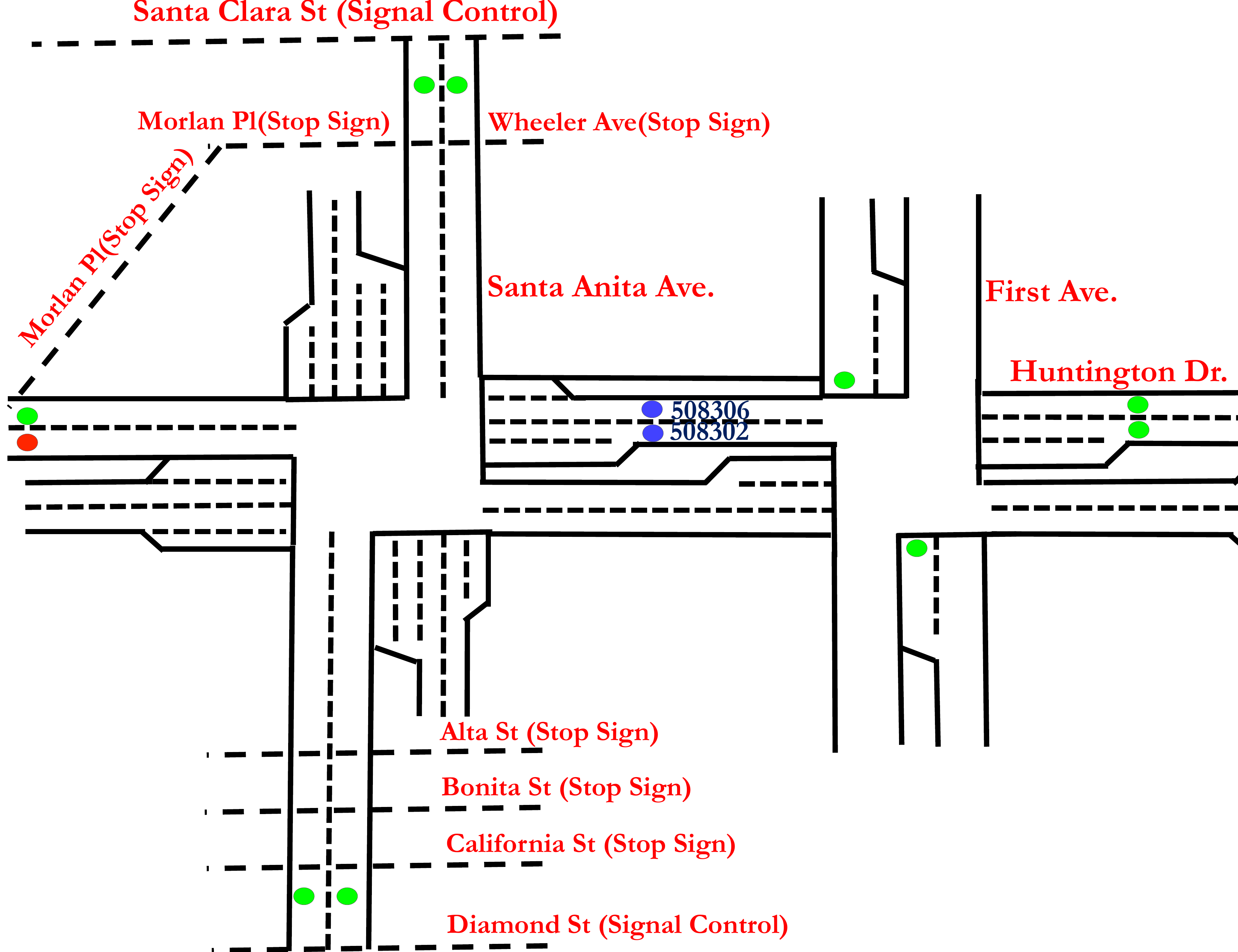}
  \caption{Detector layout at the study site in Arcadia. The detectors we examine in this study are 508302 and 508306.}
  \label{fig:studysite}
\end{figure}

The data used in this report is part of a larger dataset collected for the I-210 Connected Corridors Project.\footnote{\url{https://connected-corridors.berkeley.edu/i-210-pilot-landing-page}} The project dataset includes traffic flow data from stopbar and advance detectors, maps of the cities and sensor layouts, and the corresponding signal timing sheets. We surveyed detectors along Huntington Dr. between Santa Clara St. and Second Ave. in the city of Arcadia (Fig. \ref{fig:studysite}).

\begin{figure}
  \centering
  \begin{subfigure}{0.53\columnwidth}
    \centering
    \includegraphics[clip,trim={9.1cm 8cm 12.3cm 5cm},width=\linewidth]{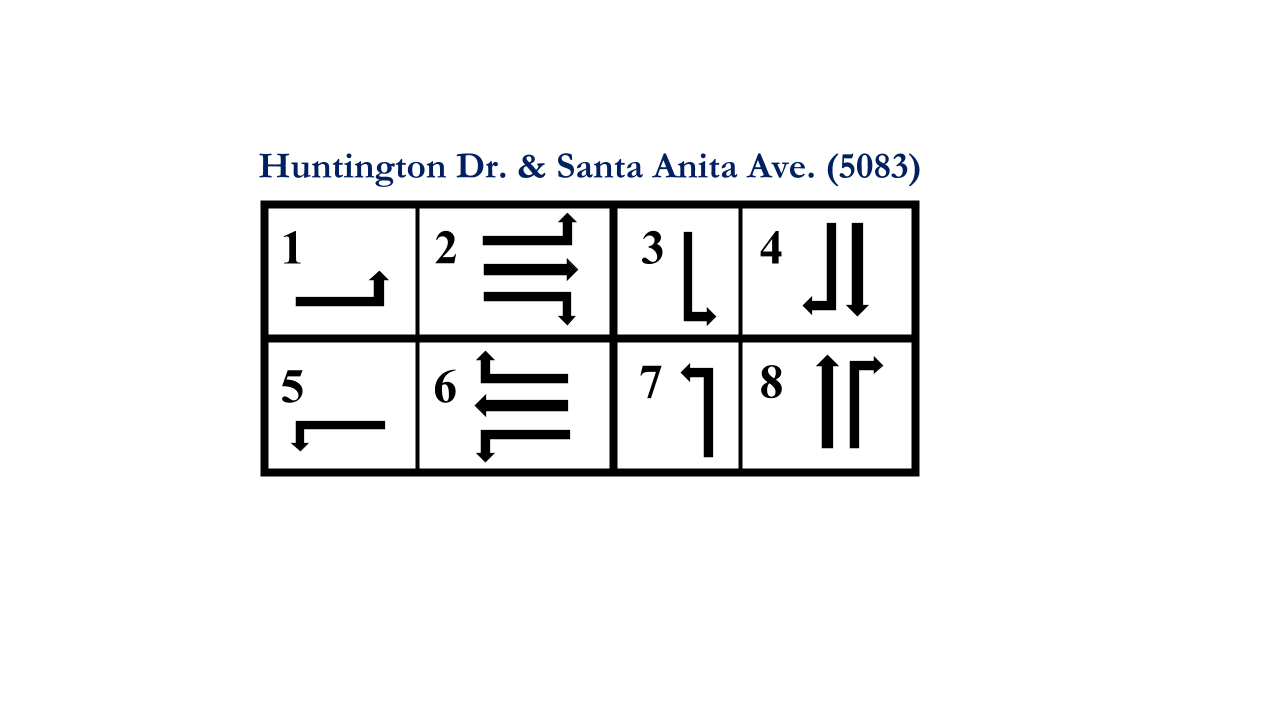}
  \end{subfigure}
  \begin{subfigure}{0.45\columnwidth}
    \centering
    \includegraphics[clip,trim={11.9cm 8cm 13.1cm 5cm},width=\linewidth]{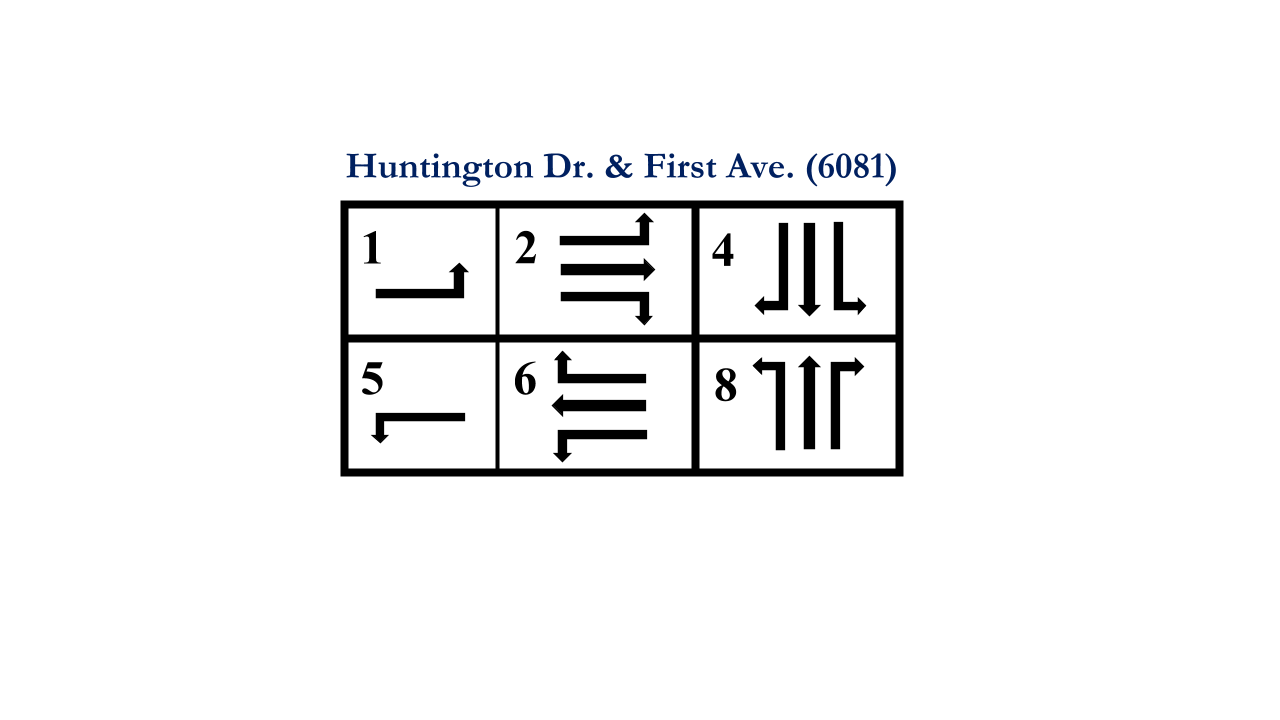}
  \end{subfigure}
  \caption{The signal phases for the upstream (6081) and downstream (5083) intersections at our study site.}
  \label{fig:phases}
\end{figure}

In particular, we focus on detectors 508302 and 508306. These detectors were selected because they are heavily covered by both advance and stopbar detectors in both the upstream and downstream directions and for Through, Right Turn, and Left Turn movements. The advance detectors for the downstream turn directions are several blocks down; while there are some leakages that prevent the system from being fully closed, they are only at minor intersections with stop signs. We call this ideal situation the Full Information scenario. See Fig. \ref{fig:phases} for the signal phase cycle and Table \ref{tab:phase_plans} for the signal timing plans.

\setlength{\tabcolsep}{2pt}
\begin{table}
	\footnotesize
	\centering
	\caption{Signal timing plans at intersection 5083 in Arcadia. "G", "Y", and "R" stand for "Green Time", "Yellow Time", and "All-Red Time", respectively. The green times provided in the table are the maximum ones from the controller settings. All values are provided in seconds.}
	\label{tab:phase_plans}
	\begin{tabular}[c]{|c|c|c|c|c|c|c|c|c|c|c|}
		\Xhline{2pt}
		\multicolumn{11}{|c|}{Huntington Dr \& Santa Anita Ave (ID: 5083)}\\
		\hline
        \multirow{3}{*}{\shortstack[c]{Plan\\Name}} & \multirow{3}{*}{\shortstack[c]{Activation\\Times}} & \multirow{3}{*}{\shortstack[c]{Cycle\\Length}} & \multicolumn{8}{c|}{Phases}\\
		\cline{4-11}
		& & &\multicolumn{2}{c|}{1 \& 5}&\multicolumn{2}{c|}{2 \& 6}&\multicolumn{2}{c|}{3 \& 7}&\multicolumn{2}{c|}{4 \& 8}\\
		\cline{4-11}
		& & &{G}&{Y+R}&{G}&{Y+R}&{G}&{Y+R}&{G}&{Y+R}\\
		\hline
		\multirow{2}{*}{$E$}&{0:00-6:00}&\multirow{2}{*}{110}&\multirow{2}{*}{20}&\multirow{2}{*}{3}&\multirow{2}{*}{27}&\multirow{2}{*}{5}&\multirow{2}{*}{20}&\multirow{2}{*}{3}&\multirow{2}{*}{27}&\multirow{2}{*}{5}\\
		{}&{21:00-24:00}&{}&{}&{}&{}&{}&{}&{}&{}&{}\\
		\hline
		\multirow{2}{*}{$P_1$}&{9:00-15:30}&\multirow{2}{*}{120}&\multirow{2}{*}{15}&\multirow{2}{*}{3}&\multirow{2}{*}{39}&\multirow{2}{*}{5}&\multirow{2}{*}{14}&\multirow{2}{*}{3}&\multirow{2}{*}{36}&\multirow{2}{*}{5}\\
		{}&{19:00-21:00}&{}&{}&{}&{}&{}&{}&{}&{}&{}\\
		\hline
		{$P_2$}&{6:00-9:00}&{120}&{11}&{3}&{46}&{5}&{11}&{3}&{36}&{5}\\
		\hline
		{$P_3$}&{15:30-19:00}&{120}&{15}&{3}&{41}&{5}&{12}&{3}&{36}&{5}\\		
		\Xhline{2pt}
	\end{tabular}
\end{table}

\subsection{Dataset}

\begin{figure}
  \centering
  \includegraphics[clip,trim={0cm 2.3cm 0cm 0cm},width=\columnwidth]{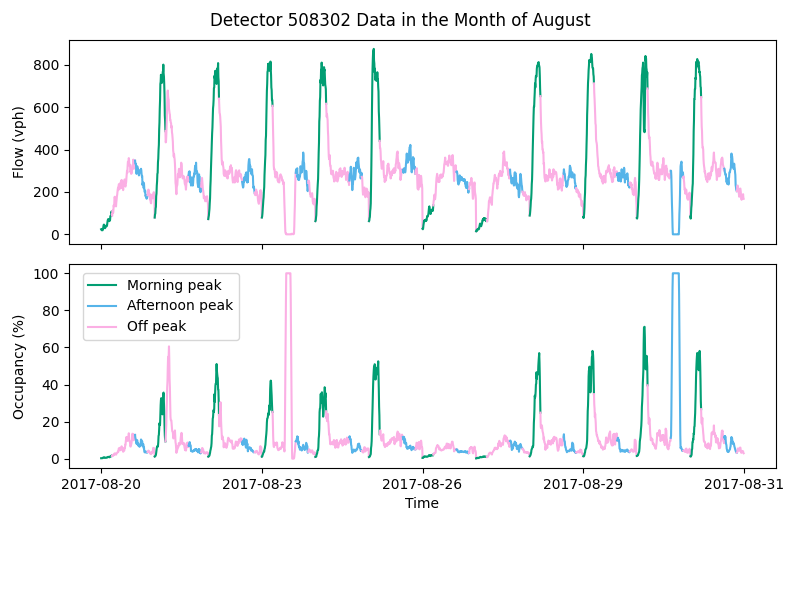}
  \caption{Flow and occupancy measured in the month of August by detector 508302.}
  \label{fig:detector_508302_data}
\end{figure}

\begin{figure}
  \centering
  \includegraphics[clip,trim={0cm 0cm 0cm 0.8cm},width=0.95\linewidth]{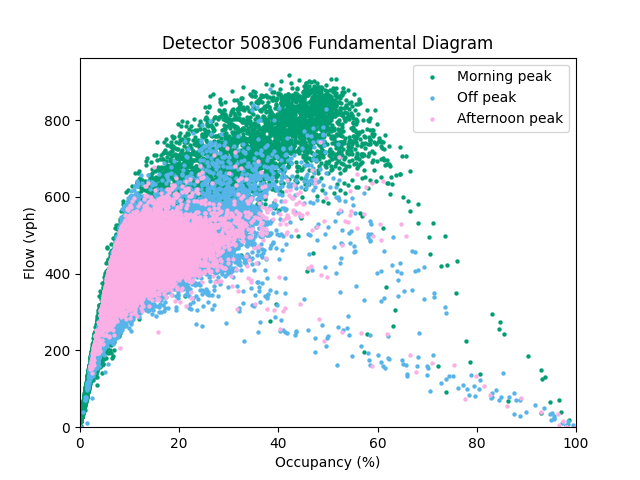}
  \caption{Flow-occupancy diagram for detector 508306, one of our detectors of interest.}
  \label{fig:fundamental_diagrams}
\end{figure}

The dataset includes flow and occupancy measurements of advance and stopbar detectors from 1/1/2017 to 12/31/2017 aggregated into five-minute intervals. Visualizations of the data from an advance detector (Fig. \ref{fig:detector_508302_data}) confirm that the measurements are highly cyclical. Flow measurements are the cleanest, whereas occupancy measurements are slightly noisier. Furthermore, data for the morning and afternoon peaks are more consistent and have larger magnitude changes than off peak data. Stopbar detectors produce noisier flow measurements and much higher occupancy values compared to advance detectors, located further upstream from the intersection. We plot a flow-occupancy graph for the data from detector 508306 (Fig. \ref{fig:fundamental_diagrams}) and note that it exhibits the trapezoidal shape that is typical of traffic fundamental diagrams. The morning peak reaches congestion and queue spillback far more often than the other two periods. Each period has its own set of signal phase timing plans, which explains the varying parameters.

Detector health and signal phase timings are collected at a granularity of one day. The signals at these intersections use four different plans: P1, P2, P3, and E, which correspond respectively to off peak, morning peak, afternoon peak, and nighttime (Table \ref{tab:phase_plans}). As expected, data from P2 and P3 have larger magnitude than data from P1 and E and exhibit very obvious cyclical patterns. The P2 and P3 plans are only active on weekdays, so we train and predict only on weekday data for the morning and afternoon peaks.

\paragraph{Preprocessing}

We performed detector health analysis to filter out spurious data. Because the DCRNN requires data for all of the detectors at each timestamp, we kept data only from days where all 12 detectors in our system were healthy. However, one downstream through detector was faulty for the entire year; as a result, we ignored it at the expense of introducing another impurity in our closed system. Otherwise, the detectors were fairly healthy, with only a few outages throughout the year, so not much data was dropped.

\section{Analysis}

\label{analysis}

For our experiments, we reused most of the original DCRNN code\footnote{Code available at \url{https://github.com/victorchan314/DCRNN}}, with some adaptations to test variations of the algorithm. The hyperparameters used in the original paper produced positive results for us. We train for 100 epochs with batch size 64, decay the learning rate by 0.1 from an initial 0.1 every 10 epochs, and use scheduled sampling with an inverse sigmoid decay function to slowly introduce predicted data as labels. We focus on Mean Absolute Percentage Error (MAPE) as a normalized metric, although we include Root Mean Squared Error (RMSE) and Mean Absolute Error (MAE) for reference. We compared the results to the following baselines: Constant Mean, Seasonal Naive/Historical Average, ARIMAX $(2, 1, 0, 1)$, and GRU. For ARIMAX, we experimented with a seasonal term and an online version, but saw no improvement over ARIMAX. Our GRU architecture follows the architecture used in \cite{DCRNN}. Like with DCRNN, two encoder cells and two decoder cells are combined in a seq2seq architecture. Thus, the only difference between DCRNN and GRU is the diffusion convolution with the incorporation of signal phase timing data.

Experiments were conducted separately on three of the traffic signal timing plans: P1, P2, and P3. The difference in traffic flow between them is significant, so we trained separate models for each plan instead of learning one model for the entire dataset. Data from the nighttime plan E is sparse and noisy; moreover, because nighttime periods exhibit little congestion, it is less useful to predict for them than for the daytime, so we did not run experiments on plan E.

All experiments predict with a horizon of six, which is equivalent to half an hour. We test four window sizes: 15 minutes (15m), half an hour (30m), one hour (1hr), and two hours (2hr), corresponding to 3, 6, 12, and 24 data points. Because the lengths of some periods during the plans were short, for example only three hours or 36 points for the morning peak, we include a \textit{start buffer} of the previous plan's data at the beginning of each period of each plan in order to have access to more data points. The length of the start buffer is equal to the window size. For example, for data from plan P2 with a 1hr window size, we include the hour of data from plan E from 5:00 to 6:00. This way, the number of data points for each of the different window sizes remains the same.

\subsection{DCRNN with Full Information}

The DCRNN was our model of choice because we had access to full information with which we could populate a transition matrix. From the signal phase timing data, we knew the weights between all pairs of detectors in the network.

\setlength{\tabcolsep}{2pt}
\begin{table*}[t]
\centering
\footnotesize
\caption{Test prediction errors for the morning peak with full information and flow only.}
\label{tab:full-information_P2}
\begin{tabular}{c|c|c|c|c|c|c|c|c|c|c|c|c|c}
& & \multicolumn{12}{|c}{Window Size} \\
\cline{3-14}
Method & Metric
& \multicolumn{3}{|c}{15 min}
& \multicolumn{3}{|c}{30 min}
& \multicolumn{3}{|c}{1 hr}
& \multicolumn{3}{|c}{2 hr} \\
\hline
\multirow{3}{*}{\shortstack[c]{DCRNN}}
% & MSE & 521.04 & 2725 & 10481 & 485.99 & 2831 & 9832 & 379.28 & 2444 & 8958 & 347.63 & 2199 & 8348 \\
& RMSE & 22.83 & 52.2 & 102.38 & 22.05 & 53.22 & 99.16 & 19.48 & 49.44 & 94.65 & 18.64 & 46.89 & 91.37 \\
& MAE & 16.33 & 34.37 & 64.03 & 15.13 & 33.97 & 60.54 & 13.9 & 31.75 & 57.18 & 13.4 & 30.67 & 55.83 \\
& MAPE & \textbf{5.38\%} & 11.33\% & \textbf{17.44\%} & 5.14\% & 11.23\% & \textbf{16.81\%} & 5.06\% & 10.52\% & \textbf{16.26\%} & 4.58\% & \textbf{9.63\%} & \textbf{15.67\%} \\
\hline
\multirow{3}{*}{\shortstack[c]{Constant\\Mean}}
% & MSE & 68980 & 65071 & 59296 & 68980 & 65071 & 59296 & 68980 & 65071 & 59296 & 68980 & 65071 & 59296 \\
& RMSE & 262.64 & 255.09 & 243.51 & 262.64 & 255.09 & 243.51 & 262.64 & 255.09 & 243.51 & 262.64 & 255.09 & 243.51 \\
& MAE & 238.82 & 230.16 & 213.45 & 238.82 & 230.16 & 213.45 & 238.82 & 230.16 & 213.45 & 238.82 & 230.16 & 213.45 \\
& MAPE & 137.6\% & 123.2\% & 105.0\% & 137.6\% & 123.2\% & 105.0\% & 137.6\% & 123.2\% & 105.0\% & 137.6\% & 123.2\% & 105.0\% \\
\hline
\multirow{3}{*}{\shortstack[c]{Seasonal\\Naive}}
% & MSE & 69022 & 65112 & 59355 & 69022 & 65112 & 59355 & 69022 & 65112 & 59355 & 69022 & 65112 & 59355 \\
& RMSE & 262.72 & 255.17 & 243.63 & 262.72 & 255.17 & 243.63 & 262.72 & 255.17 & 243.63 & 262.72 & 255.17 & 243.63 \\
& MAE & 238.77 & 230.1 & 213.44 & 238.77 & 230.1 & 213.44 & 238.77 & 230.1 & 213.44 & 238.77 & 230.1 & 213.44 \\
& MAPE & 137.5\% & 123.2\% & 105.1\% & 137.5\% & 123.2\% & 105.1\% & 137.5\% & 123.2\% & 105.1\% & 137.5\% & 123.2\% & 105.1\% \\
\hline
\multirow{3}{*}{\shortstack[c]{ARIMAX}}
% & MSE & 1006 & 6915 & 25247 & 1006 & 6915 & 25247 & 1006 & 6915 & 25247 & 1006 & 6915 & 25247 \\
& RMSE & 31.72 & 83.16 & 158.89 & 31.72 & 83.16 & 158.89 & 31.72 & 83.16 & 158.89 & 31.72 & 83.16 & 158.89 \\
& MAE & 23.93 & 63.59 & 122.44 & 23.93 & 63.59 & 122.44 & 23.93 & 63.59 & 122.44 & 23.93 & 63.59 & 122.44 \\
& MAPE & 7.3\% & 17.81\% & 28.63\% & 7.3\% & 17.81\% & 28.63\% & 7.3\% & 17.81\% & 28.63\% & 7.3\% & 17.81\% & 28.63\% \\
\hline
\multirow{3}{*}{\shortstack[c]{GRU}}
% & MSE & 578.2 & 2846 & 9373 & 466.49 & 2809 & 9668 & 385.91 & 2412 & 8968 & 348.2 & 2355 & 9414 \\
& RMSE & 24.05 & 53.35 & 96.82 & 21.6 & 53.0 & 98.33 & 19.64 & 49.11 & 94.7 & 18.66 & 48.53 & 97.03 \\
& MAE & 17.34 & 36.63 & 63.92 & 15.42 & 36.07 & 64.96 & 14.34 & 33.43 & 61.49 & 13.41 & 33.05 & 64.42 \\
& MAPE & 5.39\% & \textbf{11.1\%} & 17.46\% & \textbf{4.97\%} & \textbf{10.94\%} & 17.63\% & \textbf{4.56\%} & \textbf{10.15\%} & 16.96\% & \textbf{4.37\%} & 9.98\% & 17.5\% \\
\hline
% & & 5 min & 15 min & 1 hr & 5 min & 15 min & 1 hr & 5 min & 15 min & 1 hr & 5 min & 15 min & 1 hr \\
& & 5m & 15m & 30m & 5m & 15m & 30m & 5m & 15m & 30m & 5m & 15m & 30m \\
\cline{3-14}
& & \multicolumn{12}{c}{Horizon}
\end{tabular}
\end{table*}

\setlength{\tabcolsep}{2pt}
\begin{table*}[t]
\centering
\footnotesize
\caption{Test prediction errors for off peak with full information and flow only. Constant Mean and Seasonal Naive baselines had poor performance and are thus omitted.}
\label{tab:full-information_P1}
\begin{tabular}{c|c|c|c|c|c|c|c|c|c|c|c|c|c}
& & \multicolumn{12}{|c}{Window Size} \\
\cline{3-14}
Method & Metric
& \multicolumn{3}{|c}{15 min}
& \multicolumn{3}{|c}{30 min}
& \multicolumn{3}{|c}{1 hr}
& \multicolumn{3}{|c}{2 hr} \\
\hline
\multirow{3}{*}{\shortstack[c]{DCRNN}}
% & MSE & 289.71 & 974.75 & 1842 & 235.67 & 920.15 & 1851 & 206.54 & 794.93 & 1774 & 246.19 & 802.61 & 1721 \\
& RMSE & 17.02 & 31.22 & 42.92 & 15.35 & 30.33 & 43.03 & 14.37 & 28.19 & 42.13 & 15.69 & 28.33 & 41.49 \\
& MAE & 13.33 & 23.52 & 31.32 & 11.9 & 22.51 & 31.21 & 11.22 & 21.48 & 30.79 & 12.22 & 21.68 & 30.6 \\
& MAPE & \textbf{4.18\%} & \textbf{7.58\%} & \textbf{10.53\%} & \textbf{3.72\%} & \textbf{7.26\%} & \textbf{10.48\%} & \textbf{3.54\%} & \textbf{7.01\%} & \textbf{10.44\%} & 3.86\% & 6.94\% & \textbf{10.02\%} \\
\hline
\multirow{3}{*}{\shortstack[c]{ARIMAX}}
% & MSE & 358.68 & 1483 & 3337 & 358.68 & 1483 & 3337 & 358.68 & 1483 & 3337 & 348.53 & 1443 & 3310 \\
& RMSE & 18.94 & 38.52 & 57.77 & 18.94 & 38.52 & 57.77 & 18.94 & 38.52 & 57.77 & 18.67 & 37.99 & 57.54 \\
& MAE & 14.48 & 28.37 & 41.78 & 14.48 & 28.37 & 41.78 & 14.48 & 28.37 & 41.78 & 14.33 & 28.07 & 41.69 \\
& MAPE & 4.51\% & 8.83\% & 13.31\% & 4.51\% & 8.83\% & 13.31\% & 4.51\% & 8.83\% & 13.31\% & 4.47\% & 8.73\% & 13.28\% \\
\hline
\multirow{3}{*}{\shortstack[c]{GRU}}
% & MSE & 310.7 & 1055 & 2107 & 247.61 & 948.38 & 2039 & 220.41 & 879.1 & 2018 & 197.81 & 825.01 & 2008 \\
& RMSE & 17.63 & 32.48 & 45.9 & 15.74 & 30.8 & 45.17 & 14.85 & 29.65 & 44.92 & 14.06 & 28.72 & 44.81 \\
& MAE & 13.66 & 24.69 & 34.07 & 12.16 & 23.29 & 33.39 & 11.52 & 22.47 & 33.2 & 10.92 & 21.71 & 32.82 \\
& MAPE & 4.29\% & 7.77\% & 10.84\% & 3.82\% & 7.33\% & 10.66\% & 3.62\% & 7.07\% & 10.59\% & \textbf{3.44\%} & \textbf{6.89\%} & 10.52\% \\
\hline
% & & 5 min & 15 min & 1 hr & 5 min & 15 min & 1 hr & 5 min & 15 min & 1 hr & 5 min & 15 min & 1 hr \\
& & 5m & 15m & 30m & 5m & 15m & 30m & 5m & 15m & 30m & 5m & 15m & 30m \\
\cline{3-14}
& & \multicolumn{12}{c}{Horizon}
\end{tabular}
\end{table*}

\setlength{\tabcolsep}{2pt}
\begin{table*}[t]
\centering
\footnotesize
\caption{Test prediction errors for the afternoon peak with full information and flow only. Constant Mean and Seasonal Naive baselines had poor performance and are thus omitted.}
\label{tab:full-information_P3}
\begin{tabular}{c|c|c|c|c|c|c|c|c|c|c|c|c|c}
& & \multicolumn{12}{|c}{Window Size} \\
\cline{3-14}
Method & Metric
& \multicolumn{3}{|c}{15 min}
& \multicolumn{3}{|c}{30 min}
& \multicolumn{3}{|c}{1 hr}
& \multicolumn{3}{|c}{2 hr} \\
\hline
\multirow{3}{*}{\shortstack{DCRNN}}
% & MSE & 330.6 & 942.6 & 1591 & 268.6 & 857.35 & 1448 & 227.86 & 757.32 & 1346 & 216.88 & 733.63 & 1401 \\
& RMSE & 18.18 & 30.7 & 39.89 & 16.39 & 29.28 & 38.06 & 15.1 & 27.52 & 36.69 & 14.73 & 27.09 & 37.44 \\
& MAE & 14.51 & 24.27 & 30.94 & 13.02 & 23.17 & 29.51 & 11.98 & 21.78 & 28.41 & 11.75 & 21.41 & 28.94 \\
& MAPE & \textbf{4.01\%} & \textbf{6.69\%} & \textbf{8.44\%} & \textbf{3.62\%} & \textbf{6.37\%} & \textbf{8.09\%} & \textbf{3.3\%} & \textbf{6.0\%} & \textbf{7.82\%} & \textbf{3.26\%} & \textbf{5.94\%} & \textbf{8.0\%} \\
\hline
\multirow{3}{*}{\shortstack{ARIMAX}}
% & MSE & 375.2 & 1248 & 2587 & 375.2 & 1248 & 2587 & 375.2 & 1248 & 2587 & 375.2 & 1248 & 2587 \\
& RMSE & 19.37 & 35.34 & 50.86 & 19.37 & 35.34 & 50.86 & 19.37 & 35.34 & 50.86 & 19.37 & 35.34 & 50.86 \\
& MAE & 15.34 & 28.01 & 40.17 & 15.34 & 28.01 & 40.17 & 15.34 & 28.01 & 40.17 & 15.34 & 28.01 & 40.17 \\
& MAPE & 4.22\% & 7.76\% & 11.27\% & 4.22\% & 7.76\% & 11.27\% & 4.22\% & 7.76\% & 11.27\% & 4.22\% & 7.76\% & 11.27\% \\
\hline
\multirow{3}{*}{\shortstack[c]{GRU}}
% & MSE & 371.43 & 1205 & 2415 & 299.3 & 1163 & 2456 & 262.24 & 1090 & 2434 & 235.82 & 994.51 & 2338 \\
& RMSE & 19.27 & 34.72 & 49.15 & 17.3 & 34.11 & 49.57 & 16.19 & 33.02 & 49.34 & 15.36 & 31.54 & 48.36 \\
& MAE & 15.27 & 27.63 & 38.66 & 13.8 & 27.21 & 39.04 & 12.89 & 26.3 & 38.95 & 12.27 & 25.08 & 38.16 \\
& MAPE & 4.21\% & 7.7\% & 10.89\% & 3.8\% & 7.51\% & 10.98\% & 3.55\% & 7.25\% & 10.89\% & 3.37\% & 6.92\% & 10.7\% \\
\hline
% & & 5 min & 15 min & 1 hr & 5 min & 15 min & 1 hr & 5 min & 15 min & 1 hr & 5 min & 15 min & 1 hr \\
& & 5m & 15m & 30m & 5m & 15m & 30m & 5m & 15m & 30m & 5m & 15m & 30m \\
\cline{3-14}
& & \multicolumn{12}{c}{Horizon}
\end{tabular}
\end{table*}

\paragraph{Morning Peak}

The results for the morning peak are available in Table \ref{tab:full-information_P2}. We note several trends in DCRNN results. Prediction error increases drastically as we predict flow at longer horizons. Unsurprisingly, data points further from input data points have higher entropy. Moreover, in general, error decreases as window size increases. However, in some situations, error increases when we use a 2hr window. This seems to be a common pattern across other experiments as well. There is enough variation in the training procedure that performance saturates once window size reaches 1hr. Because using longer windows requires longer times to train the model, a practical implementation of DCRNN could use a shorter window and still achieve near-optimal performance. For this location, 1hr is the plateau for window size.

We compare DCRNN results to that of ARIMAX and GRU. ARIMAX has decent performance on the data. However, for the longer-horizon 15m and 30m predictions, DCRNN performs much better than ARIMAX. The difference is less pronounced between DCRNN and GRU. In fact, GRU performance is very close to DCRNN performance, achieving lower error in several cases, even though the difference is within the bounds of random variation. Notably, the gap narrows for 15m predictions, and for 30m predictions, DCRNN achieves lower errors.

The difference between DCRNN and the other models becomes clear when we predict for longer horizons. DCRNN is able to learn long-term temporal characteristics of the system more effectively than GRU, the next-best model, can. As a result, DCRNN consistently outperforms GRU for long horizon predictions, even when GRU achieves lower error for the 5m and 15m predictions. Moreover, whereas GRU error plateaus at the 1hr window size, DCRNN error continues to drop when we use a 2hr window size, indicating that with more training data, we would see even better predictions from DCRNN. Because traffic during the morning peak is very irregular and complicated, the model requires more examples to learn the rich patterns present in the data.

\begin{figure}[t]
  \centering
  \includegraphics[clip,trim={1.9cm 0cm 1.3cm 0.6cm},width=\columnwidth]{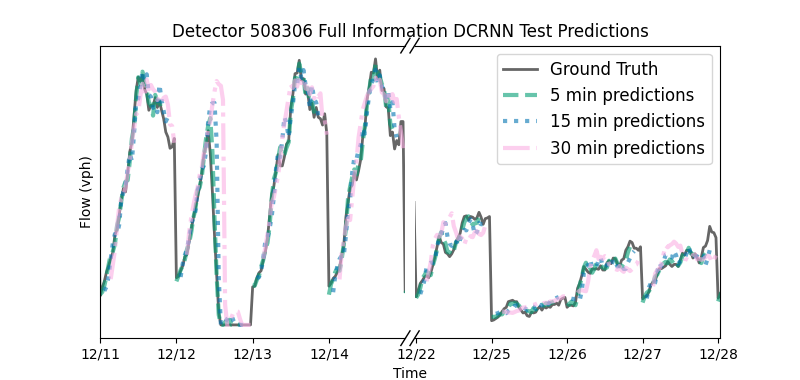}
  \caption{Test predictions of DCRNN on flow measured in the month of December by detector 508306. The model was trained with a window size of 1hr and a horizon of 6 data points, or half an hour.}
  \label{fig:detector_508306_predictions}
\end{figure}

The comparison between DCRNN predictions and the ground truth is illustrated in Fig. \ref{fig:detector_508306_predictions}. All three horizons result in predictions that are very close to the ground truth, although the half-hour predictions are clearly less accurate. We note a consistent overshooting problem in the graph. When the ground truth data maintains its cyclical course, even with sawtooth edges, the predictions match very closely. However, for irregular changes in shape, the model's predictions at each time step continue along the previous trajectory and diverge from the ground truth. For example, on 12/12, traffic grinds to a halt in the middle of the morning peak, likely due to an exogenous event such as an accident. However, the model's predictions overshoot this drop and continue upwards in the direction that traffic flow would usually travel. After seeing the atypically low measurements, the model tries to correct by predicting a very sudden sharp downward trajectory. At this point, the output of DCRNN is actually negative, overshooting the flattening out of flow, so we clip the value to zero when predicting. While DCRNN is able to learn the cyclical patterns, it struggles to adapt to outliers and rare exogenous events.

\begin{figure}[t]
  \centering
  \includegraphics[clip,trim={1.9cm 0cm 1.3cm 0.6cm},width=\columnwidth]{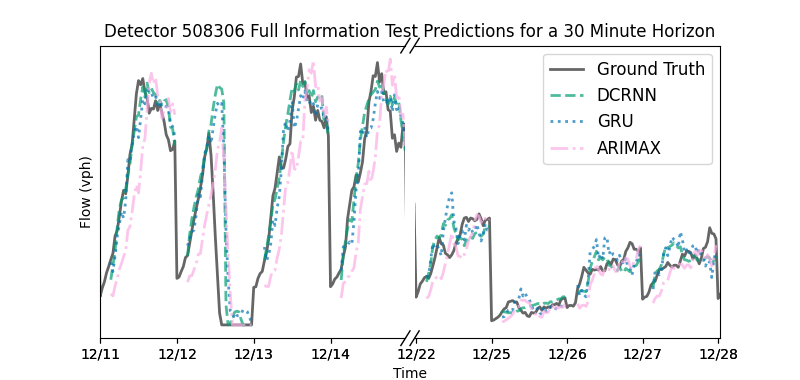}
  \caption{Test predictions of DCRNN, GRU, and ARIMAX on flow measured in the month of December by detector 508306. The model was trained with a window size of 1hr and a horizon of 6 data points, or half an hour. Here we show the predictions with a horizon of 6, where DCRNN improves the most compared to the other methods.}
  \label{fig:detector_508306_baselines}
\end{figure}

In Fig. \ref{fig:detector_508306_baselines}, we see the comparison between DCRNN and its two closest baselines for half-hour predictions. ARIMAX exhibits the cyclical shape of the ground truth data, but lags behind by several time steps, and thus is unable to predict the future accurately. GRU is similar to DCRNN, except that it tends to not follow the ground truth pattern as closely as does DCRNN. Especially during the congested morning periods, DCRNN reaches the same amplitude that the ground truth reaches, but GRU peaks too soon and turns downwards before the ground truth flow subsides. Even during winter vacation, from Christmas (12/25) to New Year's Eve (12/31), when traffic flow has quieted down, DCRNN adapts to the smaller peaks more quickly than GRU does.

\paragraph{Other Times}

In both the afternoon peak (Table \ref{tab:full-information_P3}) and off peak (Table \ref{tab:full-information_P1}) results, we recognize many of the same trends present in the morning peak experiments. Longer windows consistently produce lower prediction errors, but not past 1hr. For the off peak plan, we see the same pattern as with the morning peak. DCRNN outperforms all of the other models except for with a 2hr window, where GRU achieves lower error for 5m and 15m horizon predictions. However, for the afternoon peak plan, DCRNN outperforms every other model. Thus, the results substantiate the hypothesis that a major advantage of DCRNN is its ability to learn the pattern of traffic flow for long horizon predictions. The ARIMAX and GRU models are able to memorize the gist of the data trends, but fail to understand finer details of the data.

One notable difference between these two periods and the morning peak is the drastically lower prediction error, often dipping below four percent, even for the baselines. The magnitude of data from outside the morning peak is much lower, and therefore the peaks are not as pronounced. Because there is less up-and-down variation in the data, the trends are easier for the models to learn and predict. Moreover, there is less contrast between different days of the week. Traffic during the afternoon peak is simpler than during the morning peak, so even with a limited amount of data, DCRNN enjoys the full benefits of the signal phase timing plans.

For the afternoon peak, DCRNN still has a clear improvement over GRU, especially for the 30m predictions. However, for the off peak plan, although DCRNN consistently achieves lower error than GRU, the decrease is far smaller. We believe that this is due to the fact that signal timing plans have the most substantial effect on traffic flow when there is congestion but not queue spillback. During these peak hours, cars are fully subject to the signal phases. Thus, the transition matrix representing the diffusion process comes into play. During the off peak hours, there is little congestion, and therefore little benefit from modeling the intersection.

\paragraph{Flow and Occupancy}

In theory, if we provide additional sensor information to the model, it should perform at least as well as before. This expectation was realized in experiments for the off peak and afternoon peak periods, where the results did not change significantly when occupancy was included in the input data. However, when we included occupancy for the morning peak, the error increased by a not insignificant amount. The MAPE for 30m predictions regressed by one percent for DCRNN and two percent for GRU. Notably, we saw the same trend as before, where DCRNN always outperformed the other baselines for 30m predictions, but was occasionally outperformed by GRU for 5m and 15m predictions. However, DCRNN and GRU both suffered when occupancy was added as an extra feature. We do not fully understand why this phenomenon occurs; however, our hypothesis is that occupancy data from the detectors is noisy enough that it affects the long-term relationships that the models learn.

\subsection{Incomplete Information Scenarios}

We chose a specific group of detectors for our experiment because they represented a system that was very close to being fully-observed for both upstream and downstream lanes. However, cities may not have the resources to cover every lane on every road with detectors, and even then, real-world detectors occasionally fail, as exemplified by our data. Even in our network, detectors 508302 and 508306 are the only two detectors with extensive coverage, and the system is still not fully closed. In order to test situations in which full data is not available, we ran experiments with augmented data. We provide a brief summary here; for complete results, see \cite{Chan:EECS-2020-68}.

The first scenario is the case of incomplete information, when we do not have full coverage. We simulated this by creating a new transition matrix with only a subset of the detectors and predicting using data from only that subset. Omitting upstream detectors resulted in saturation of prediction accuracy for longer horizons, indicating that the full information scenario might have superior performance with more data. Omitting downstream detectors actually produced a consistent slight improvement for the morning peak. This phenomenon is likely caused by the imperfect closed system in the downstream direction, with an unhealthy detector and multiple stop sign intersections violating the diffusion assumptions of the transition matrix. Overall, however, these differences were not significant.

The second scenario is the case of unhealthy detector data. We used the same transition matrix as the full information case, but zeroed out part of the data to simulate a situation in which the number of detectors is fixed, but some of the detectors do not provide good data during training. The augmented data was evaluated in two ways to determine robustness to unreliable input: predicting on the data with trained models from the full information scenario, and training new models.

In general, using the full information models to predict with the presence of any unhealthy detectors greatly diminished accuracy. Retraining on the augmented data alleviated the error, although it was unable to close the gap. Clearly, the full information models rely on data from all detectors for the most accurate predictions. Surprisingly, even when all but the two noisy stopbar detectors were augmented, short horizon predictions were not affected after retraining. While flow from our two detectors of interest is sufficient for one-step predictions, flow from the other detectors is crucial for long horizon predictions.

Our final scenario investigated setting the detector data to zero on a certain proportion of randomly-selected days to simulate temporary outages. We analyzed four percentages of days to augment: 5\%, 10\%, 25\%, and 50\%. For all plans and horizons, using the full information matrix to predict significantly curtailed performance, with MAPE surging to past 50\% as a larger portion of the data was augmented. Even after retraining, the errors were large and exhibited high variation. We can conclude that data quality is of utmost importance to our model. Training data must be carefully preprocessed to avoid detrimental effects. Unlike in the previous scenarios, where the model is robust to misbehaving detectors, here the quality of the data itself is degraded. While the model is able to absorb some of the impact and produce decent results in some cases, it produces much more accurate and consistent results when no data is corrupt.

\subsection{Other Experiments}

We also tested several DCRNN setup variations to investigate whether these alterations would generate more accurate predictions. First, we trained a different DCRNN for each day of the week, as \cite{lowrank} discovered significant changes in the traffic profile between each day of the week. Second, we trained six DCRNNs to make single-horizon predictions instead of predicting all six horizons at once. Neither of these experiments improved upon the original model; splitting by day of the week resulted in similar performance, while single-horizon predictions actually performed worse. Instead of just jumping ahead without intermediate context, the model requires all of the training labels to identify temporal patterns. The original DCRNN model has the structure and expressiveness to represent traffic in our system without these extra modifications.

\section{Conclusion}

\label{conclusion}

Arterial traffic prediction is far more challenging compared to freeway traffic prediction. Spatial information plays a much more salient role and must be effectively applied to optimize prediction accuracy. In this study, we explored using signal phase timing data to generate a weighted adjacency matrix based on traffic signal phase splits. Combined with our graph convolutional model of choice, the DCRNN, we show that the signal phase timing data enhances arterial flow predictions, especially long horizon forecasts. We achieve MAPE as low as 16\% for a prediction horizon of 30 minutes for morning peak congestion. For afternoon peak and off peak data, we achieve MAPE lower than 8\% and 10\% for the same horizon. Signal phase timing data defines the relationships between detectors in the network and allows the model to learn long-term temporal relationships for long horizon predictions.

In addition, we tested numerous variations of the measurements and the detector network to investigate the effects of detector coverage and data quality on prediction performance. One surprising discovery is that detector coverage is overshadowed by detector proximity and precise measurements; as a result, we saw no significant effects after omitting stopbar detectors and distant detectors. In every scenario with simulated unhealthy data, prediction accuracy and consistency diminished, but retraining the model mitigated that decline. Short horizon predictions were not particularly affected, but long-horizon prediction error skyrocketed with just one or two unhealthy detectors. When presented with more information, our model makes good use of it to generate excellent predictions, but it is also robust to faulty detectors during training. However, at least some of the detectors must be relatively reliable---errors soared even when only 5\% of days were zeroed out. Although the data can include some anomalies, it must be relatively consistent throughout the entire dataset.

In the future, we can study extensions and variations of this work. We can train deeper and more expressive models to better learn complex patterns. The area of deep unsupervised learning is burgeoning, and because traffic network matrices are polynomial with respect to the number of detectors and the size of the graph, it would be very useful to find a compressed feature representation for the entire network state. This would be particularly beneficial for the signal phase timing data. DCRNN applies a static transition matrix, so we used planned phase splits; however, traffic plans are dynamic and reactive to traffic conditions, so the actual phase splits are different for each point in time. With a latent embedding, we could encode the signal phases for each data point instead of aggregating them into a single static matrix. Some newer graph convolutional architectures, such as Graph WaveNet \cite{Graph_WaveNet}, allow adaptive filters, so they can be applied to the problem as well.

Another prospective direction is to include even more varied types of information, such as pedestrian activity at intersections. In addition, DCRNN allows prediction of all detectors at once. We could examine flow forecasts for an entire network of sensors, even one that isn't a closed system. Flow predictions can also be applied to signal control applications to determine the effect of forecasts on travel time and queue length on urban roads. Arterial traffic predictions have many applications, so we must leverage all the data and technology in our toolbox to tackle the challenge.

% or


% \nocite{*}

\section*{Acknowledgment}

The authors express their thanks to Damian Dailisan, Umang Sharaf, Keith Anshilo Diaz, and Carissa Santos for providing thoughtful insights into the experiments.

\ifCLASSOPTIONcaptionsoff
  \newpage
\fi



\bibliographystyle{IEEEtran}
\bibliography{references/references,references/deeplearning,references/graphconv,references/applications}


\begin{IEEEbiographynophoto}{Victor Chan}
received his B.S. and M.S. in Electrical Engineering and Computer Science from the University of California, Berkeley in 2019 and 2020, respectively. His areas of expertise include deep learning and computer vision, especially in the realm of transportation systems. 
\end{IEEEbiographynophoto}

\begin{IEEEbiographynophoto}{Qijian Gan}
received the B.E. degree in automatic control from the University of Science and Technology of China, Hefei, China, in 2009, and the M.S. and Ph.D. degrees in civil engineering from University of California at Irvine, Irvine, CA, USA, in 2010 and 2014, respectively. He is currently a Research and Development Engineer in the PATH Program at the University of California at Berkeley, Berkeley, CA, USA. His main expertise includes network traffic flow theory, network modeling and simulation, traffic signal control, and data analysis.
\end{IEEEbiographynophoto}

%\newpage

\begin{IEEEbiographynophoto}{Alexandre Bayen}
received the Engineering degree in applied mathematics from Ecole Polytechnique, France, in 1998, and the M.S. degree in aeronautics and astronautics and the Ph.D. degree in aeronautics and astronautics from Stanford University in 1999 and 2003, respectively. He was a Visiting Researcher with the NASA Ames Research Center, from 2000 to 2003. In 2004, he was the Research Director of the Autonomous Navigation Laboratory, Laboratoire de Recherches Balistiques et Aerodynamiques, Ministere de la Defense, France, where he holds the rank of Major. Since 2014, he has been the Director of the Institute for Transportation Studies, where he is currently an Associate Chancellor Professor.
\end{IEEEbiographynophoto}


%\vfill


\end{document}